# Knowledge-enhanced Neural Machine Reasoning: A Review

TANMOY CHOWDHURY*, Goerge Mason University, USA
CHEN LING*, Emory University, USA
XUCHAO ZHANG, Microsoft, USA
XUJIANG ZHAO, NEC Laboratories America, USA
GUANGJI BAI, Emory University, USA
JIAN PEI, Duke University, USA
HAIFENG CHEN, NEC Laboratories America, USA
LIANG ZHAO, Emory University, USA

Knowledge-enhanced neural machine reasoning has garnered significant attention as a cutting-edge yet challenging research area with numerous practical applications. Over the past few years, plenty of studies have leveraged various forms of external knowledge to augment the reasoning capabilities of deep models, tackling challenges such as effective knowledge integration, implicit knowledge mining, and problems of tractability and optimization. However, there is a dearth of a comprehensive technical review of the existing knowledge-enhanced reasoning techniques across the diverse range of application domains. This survey provides an in-depth examination of recent advancements in the field, introducing a novel taxonomy that categorizes existing knowledge-enhanced methods into two primary categories and four subcategories. We systematically discuss these methods and highlight their correlations, strengths, and limitations. Finally, we elucidate the current application domains and provide insight into promising prospects for future research.

## 1 INTRODUCTION

Since the term *Artificial Intelligence* (AI) was coined in the last century, the ultimate research goal of AI is making an intelligent system that can reason like humans, i.e., perceive existing evidence, synthesize problem-relevant information, and infer novel knowledge to solve unseen tasks. Earlier research on AI circulated on hand-crafted and logical rules, which resulted in a strong reasoning ability but poor system scalability and generalizability. With the development of deep neural networks, state-of-the-art deep learning models excel at extracting predictive patterns for specific tasks from a massive amount of data. To further move toward *Artificial General Intelligence* (AGI), we realized that the current learning system needs to be able to think more deeply and adaptively in different tasks. Specifically, two major trends in AI research have been evidenced in recent years: 1) Going beyond pattern recognition to logical reasoning, which echos human's complementary learning strategies (i.e., *Thinking System I & II* [25]); and 2) Moving from task-specific models to general intelligence, which calls for the capability of leveraging external knowledge [13] (or even world models [19]).

In fact, human-like intelligence indeed requires the synergy between the aforementioned two trends. To obtain stronger decision-making and logical reasoning capabilities, human beings need to understand and embed logical processes of existing evidence. On the other hand, it is also essential for humans to make predictions or construct approaches toward an arbitrary (unseen) task with the help of existing knowledge. Despite the above strong motivation, the explosively-fast growing efforts on synergizing deep reasoning and knowledge augmentation are observed until very recently, thanks to the advancement of several key areas these years. Specifically, the fast-advancing high-performance computing infrastructure enables to construct extremely large neural networks (e.g., GPT-3 [6] with 175 billion parameters) that can inherently exhibit reasoning ability through implicit knowledge stored in their parameters. In addition, the recent development of various organized and structured knowledge bases can also provide abundant external information with both general and domain-specific knowledge required for reasoning.

Endowing deep model reasoning ability is challenging, yet external knowledge may provide a head start. However, using external knowledge for making hard logical inferences makes the system fragile as it depends solely on deduction. On the other hand, inductive reasoning is the core focus of traditional deep learning. Therefore, it is an open research problem to develop a *'work in tandem'* technique for both methodologies, which brings several challenges including but not limited to: **1) Difficulties in Knowledge Integration**. Explicit knowledge, especially rule-based knowledge is resistant to generalization as it can be hand-crafted or very domain specific. On

---

*Both authors contributed equally to this research.

Authors' addresses: Tanmoy Chowdhury, tchowdh6@gmu.edu, Goerge Mason University, Fairfax, Virginia, USA; Chen Ling, chen.ling@emory.edu, Emory University, Atlanta, Georgia, USA; Xuchao Zhang, xuchaozhang@microsoft.com, Microsoft, USA; Xujiang Zhao, xuzhao@nec-labs.com, NEC Laboratories America, USA; Guangji Bai, guangji.bai@emory.edu, Emory University, Georgia, USA; Jian Pei, j.pei@duke.edu, Duke University, North Carolina, USA; Haifeng Chen, haifeng@nec-labs.com, NEC Laboratories America, USA; Liang Zhao, liang.zhao@emory.edu, Emory University, Georgia, USA.

the other hand, the formation of rule structures demands different types of integration (e.g. geometric embedding). The facts gathered from the structured external knowledge (e.g. knowledge graph) can be sparse which can degrade the inference performance. Moreover, each domain data has its own hidden characteristics. So, the external knowledge needs to be domain adaptive for proper integration. During integration, determining the weight of the external knowledge raises issues that are very task specific. External knowledge faces the problem of scarcity due to inadequate valid logic, poor semantics, and insufficient labeled examples. It can also be embellished with hidden misinformation which hinders the process of extracting knowledge. Moreover, generalized external knowledge brings the challenges of systematic detection and a comprehensive understanding of subjective and domain knowledge. **2) Issues in Bridging the External Knowledge and Deep Models**. In order to combine external knowledge and deep models, we have to fill the gap between the discrete nature of knowledge rules and the continuous nature of deep models. This motivates two potential directions of strategies: one way is to make the deep model to be able to query the external knowledge. But it will meet the retrieval challenges. To reduce the effort and computation resources, concerns will arise related to process parallelism, the poor performance of mini-batch training, reuse of memory and etc. The other way is to represent the knowledge as embedding or model parameters. However, the challenge is that the pre-trained parametric models may introduce bias due to the data used to train those. Additionally, utilizing a bigger model requires a lot of computer power. **3) Challenges in Tractability and Optimization**. Logical inferences in deep learning involve common approaches like using rules as additional constraints. But using rules as constraints induces non-linear and non-convex constraint issues for parameter-oriented models. Another common approach is to use probabilistic models. But sampling rules (e.g. learning generalized multi-hop rule set) is a common trait in all approaches. As the entire number of ground rules in the real world is intractable, the issue of tractability is a common difficulty for every problem formulation. Moreover, developing an end-to-end differentiable framework is a very challenging task which further leads to the challenges of optimization.

Recently, a considerable amount of research [34, 15, 64, 73, 63] has been devoted to developing reasoning techniques with the aid of various knowledge sources in order to address the challenges mentioned above. There are enormous research areas that can benefit from knowledge-enhanced reasoning techniques, ranging from commonsense reasoning in the development of conversational AI to neuro-symbolic reasoning in constructing logic programming systems. However, most existing methods are tailored for specific application domains, but their approaches could be general enough to handle similar reasoning problems in other applications. Moreover, the developed techniques for one real-world application could potentially benefit the technical developments for another application. However, it is challenging to cross-reference these techniques across different application domains serving totally different communities. To date, we have witnessed several surveys [22, 18, 8] discussing deep reasoning techniques specialized in certain application domains. Unfortunately, we still miss a timely technical overview of very recent knowledge-enhanced deep reasoning techniques across different knowledge sources and vast application domains. This absence of a systematic summary and taxonomy in knowledge-enhanced deep reasoning techniques causes major difficulties for relevant researchers to have clear information on the existing research challenges, open problems, and vast future research directions.

To overcome these obstacles and facilitate the development of AGI, we provide the *first* comprehensive overview of current works (published after 2020) on this fast-evolving topic, knowledge-enhanced neural machine reasoning. The major contributions of this survey are summarized as follows:

- **A first systematic taxonomy of existing knowledge-enhanced deep reasoning techniques.** We categorize existing techniques based on the types of external knowledge and elucidate their formal problem definition. We summarize the relationship and pros & cons among different approaches along with details of the techniques under each subcategory.
- **A comprehensive categorization and summarization of major application domains.** We summarize the vast application domains of existing knowledge-enhanced neural machine reasoning techniques. The categorization of application domains can be easily mapped to the proposed technique taxonomy for researchers to cross-reference different application domains.
- **An insightful discussion of the current state of knowledge enhanced deep reasoning and its future trends.** Based on the summarization of existing techniques on leveraging external knowledge to aid deep reasoning, we outline an overall picture and the shape of the current research frontiers on knowledge-enhanced deep reasoning. The potential fruitful future research directions are also discussed.

## 2 PROBLEM FORMULATION AND TAXONOMY

### 2.1 Problem Formulation

**Knowledge-enhanced Deep Reasoning.** Knowledge-enhanced deep reasoning aims to manipulate previously acquired intelligence and knowledge from the training domain to execute novel tasks in unseen test domains by mining and integrating relevant knowledge from external databases.

Given a deep reasoning model $\mathcal{F}(\cdot)$ that is trained on a data set $X$ and parameterized by $\theta$, where $X$ as input data can follow sensorial input patterns (image, video, text, ...) and $\theta$ is the set of model parameters (e.g., $\theta$ can be a weight matrix if $\mathcal{F}(\cdot)$ is a multilayer perceptron). By presenting $\mathcal{F}(\cdot)$ a novel reasoning task with data $\hat{X}$, $\mathcal{F}(\cdot)$ should locate the most relevant and condensed knowledge $k$ from an external knowledge base $K$ ($k \in K, |k| \ll |K|$) to make the most accurate prediction $\hat{Y}$ such that $\hat{Y} = \mathcal{F}(\hat{X}; k, \theta)$. For example, given a commonsense reasoning task in the natural language understanding domain, if we leverage knowledge graph as the external knowledge source $K$, then we need to locate several relational paths or a subgraph as $k$ according to the commonsense question $\hat{X}$. Note that $k$ can be manually added as auxiliary training instances to better enhance the reasoning capability of $\mathcal{F}(\cdot)$ at the training stage.

## 2.2 Taxonomy

We proposed a taxonomy to show the knowledge utilization in different reasoning techniques in Figure 1. For the knowledge-enhanced reasoning techniques, we focused on 1) the type of knowledge stored in data, 2) the formation of knowledge, and 3) Knowledge Utilization. Based on these three criteria we constructed the three levels of the taxonomy.

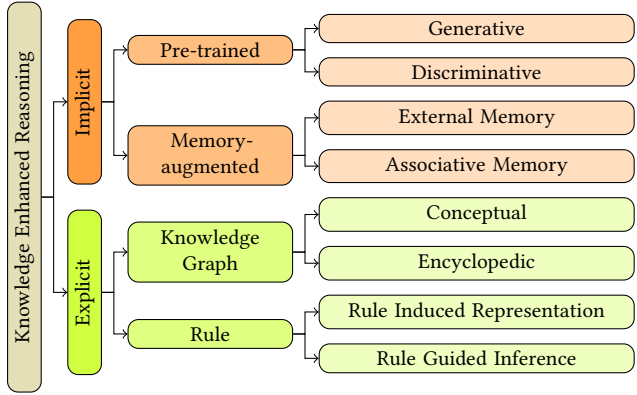

Fig. 1. The taxonomy for knowledge enhanced reasoning

The explicitness of the knowledge depends on the organization, structure, interpretation, accessibility, and real-world representation. These properties can greatly benefit the techniques used for reasoning tasks. The various types of external knowledge motivate us to divide the taxonomy mainly into two classes: 1) Explicit Knowledge and 2) Implicit Knowledge. We discussed both groups of techniques in Section 3 and Section 4 in detail, respectively.

## 3 REASONING WITH IMPLICIT KNOWLEDGE

In this section, we review the recent progress in leveraging implicit knowledge to solve reasoning tasks, where relevant knowledge is implicitly expressed in basically two forms: i.e., large-scale pre-trained models and memory-augmented neural networks. Both forms have a dependency on parameters for latent knowledge. The pre-trained model must be imported in its entirety due to its exclusive parameter dependency while the semi-parametric memory in memory-augmented neural networks can operate from the encoded index of knowledge.

## 3.1 Pre-trained Models Knowledge

With the revolutionary development of pre-training, large language models (LMs) have made significant progress in natural language processing. There is an observation that pre-trained language models may exhibit reasoning abilities if the model sizes are sufficiently large [6], and scaling up the size of LMs has already achieved success in various reasoning tasks, such as arithmetic [58], commonsense [34], and visual-symbolic [65]. As one of the most powerful pre-trained LMs, many hypes claim that the writing and reasoning capability of GPT-3 [6] are almost near-human levels. Among all techniques that leverage LMs in reasoning tasks, there are primarily two ways of mining relevant yet implicit knowledge from the model parameters.

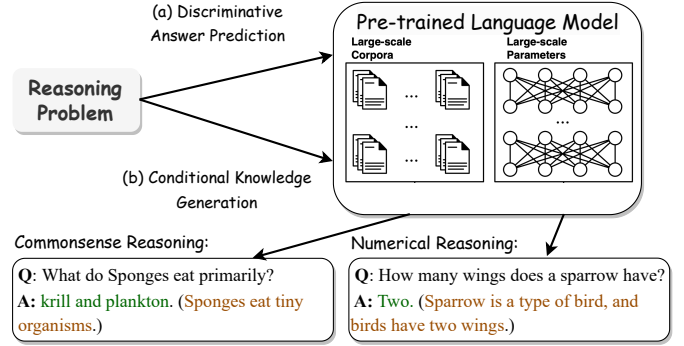

Fig. 2. Two primary paradigms of eliciting implicit knowledge from language models: (a) Discriminative Answer Prediction, and (b) Conditional Knowledge Generation.

*3.1.1 Discriminative Answer Prediction.* It has been becoming apparent that LMs themselves can be used as a tool for text understanding by formulating queries in the form of natural language questions and either directly generating textual answers [6] or picking the most likely answer from a set of answers candidates [64]. The success of pre-trained LMs is often attributed to (in-context) few-shot or zero-shot learning, namely prompt learning. A standard scheme of prompt learning requires a reasoning question $Q$, a prompt $T$, and a parameterized probabilistic LM, $p_{LM}$, the goal is to maximize the likelihood of the prediction $A$ as:

$$P(A|T,Q) = \prod_{i=1}^{|A|} p_{LM}(a_i|T, Q, a_{<i}), \quad (1)$$

where $a_i$ is $i$-th token of $A$, and $|A|$ denotes the length of $A$. By leveraging the way of prompt, LMs can either generate the continuation of a prefix (e.g., *"The winner of 2022 Nobel Prize in physics is "*) or predicts missing words [X] in a cloze-based template (e.g., "Birds have [X] wings."). We then review different types of prompts as follows.

**Single-stage Prompts.** Earlier work [43] employed template-based prompts in commonsense reasoning, which have achieved great success compared to LMs that only incorporated text generation methods. Subsequent research [68] has been conducted to demonstrate how different designs of prompts and few-shot exemplars can help the reasoning performance with in-context learning. More surprisingly, [27] even demonstrated that LMs can be zero-shot reasoners without any exemplars. By only concatenating *Let's think step by step*, LMs can consciously generate reasoning steps.

**Chain of Prompts.** Chain of Thoughts/Prompts [58] represents a more intuitive solution that decomposes a complex reasoning problem into simpler sub-problems and reasons stage by stage, which is more intuitive since the whole reasoning process cannot usually be finished in one stroke. Since its introduction, there have emerged several variants based on the chain of prompts. [50] presented a chain-of-thought way to tackle arithmetic reasoning problems, and [7] has applied chain-of-thoughts to tabular reasoning and achieved state-of-the-art performance by providing only one exemplar.

*3.1.2 Conditional Knowledge Generation.* It has been proved LMs can effectively capture world knowledge with prompts. On the other line of work, many researchers have tried to probe implicit knowledge from pre-trained LMs via conditional knowledge generation in the past few years [24] to solve the potential drawbacks (e.g., insufficient coverage and precision of the knowledge) of solely using implicit knowledge from LMs as the knowledge provider. Compared to directly leveraging prompts to make predictions, they [51, 34] additionally generate knowledge statements from pre-trained LMs, and they integrate the knowledge with the original prompt $T$ to attain a better knowledge awareness. The general paradigm is shown as follows:

$$P(A|T,Q) = \prod_{i=1}^{|A|} p_{\text{LM}}(a_i|T,Q,k_j,a_{<i}),$$
$$k_j \in K, \ K = \{k : k \sim p_{\text{LM}}(k|Q)\},$$

where $K$ is a set of knowledge statements sampled from $p_{\text{LM}}$ conditioned on the question $Q$, and $k_j$ is selected from $K$ and coupled with the designed prompt $T$ to serve as the supporting knowledge. For example, given a question: *"She was always helping at the senior center, it brought her what?"*, LMs are expected to generate a set of knowledge statements, such as {*"Some people delight in helping others.", "You'll find joy in helping others.", "Senior center is a place where everyone needs help.", ...*}. Other than querying from the LMs, there are methods [51] trying to discover additional background information in an external knowledge base for better knowledge coverage. We will discuss more details of leveraging explicit knowledge to aid reasoning tasks in Section 4.

## 3.2 Memory-augmented Knowledge

A general purpose of advanced intelligent systems should be able to maintain a memory that can store and retrieve information to handle various reasoning tasks. Despite the success of deep learning and various (pre-trained) neural network architectures, there are still a set of complex tasks that can be challenging for conventional neural networks. Those tasks often require a neural network to be equipped with an external memory in which a larger, potentially unbounded, set of facts need to be stored.

*3.2.1 External Memory.* Reasoning and inference require process as well as memory skills by humans. Neural networks are able to process tasks like image recognition (better than humans) but memory aspects are still limited (by attention mechanism, size, etc). One solution is to equip a neural network with an *external memory* such that the learning and memory functionalities are disentangled (like CPU and RAM in modern computers) and thus enjoy better capacity. A representative work of memory-augmented neural networks is Neural Turing Machine (NTM) [17], which is composed of a neural network controller and a memory bank. The controller interacts with the outside environment using input/output tensors. Unlike other neural networks, NTM can also interact with a memory matrix using attention-based read & write operators (called heads.) Every component in NTM is differentiable so the framework can be trained via gradient descent. Thanks to the attention mechanism, NTM can selectively read from or write to the memory, i.e., infer using only a small part of the knowledge in the large memory. Moreover, the memory bank in NTM allows previously learned facts to be stored in a form of implicit knowledge in the memory. Altogether,

NTM could make multiple computational steps in answering a question like RNNs while having a much stronger capacity in modeling long-term dependencies in sequential data. Many works consider improving upon NTM such as [16] where a forgetting mechanism is introduced in the dynamic memory. [44] considers leveraging the memory-augmented neural network for relational reasoning. [41] applies the memory-augmented neural network for neural symbolic reasoning.

*3.2.2 Associative Memory.* Associative memory is a type of memory that allows the recall of data based on the degree of similarity between the input and the patterns stored in memory. It refers to a memory organization in which the memory is accessed by its content as opposed to an explicit address like in the traditional computer memory system. Therefore, this type of memory allows the recall of information based on partial knowledge of its contents. [3] is the earliest work that considers associative memory for symbolic reasoning and achieved high computational efficiency. Recent works [49, 47] consider leveraging Hopfield Network as associative memory and have shown strong performance on various reasoning tasks.

## 4 REASONING WITH EXPLICIT KNOWLEDGE

Explicit or expressive knowledge is information that can be effortlessly recorded, accessed, and interpreted. The nature of explicit knowledge is logical, objective, and structured. In the last few decades, people have been extensively mining structured knowledge to support various reasoning-related tasks from explicit knowledge bases, including knowledge graphs and knowledge rules. The ability of these two types of knowledge is to clearly define the relationships between distinct entities has led to a growing common interest. The graph is heavily utilized at the input level of the model or is used to gather concepts from the raw data. But the rule has the flexibility to be used either at the representation level or inference level. In this section, we review recent techniques for utilizing different explicit knowledge forms.

### 4.1 Graph Structured Knowledge

Knowledge graphs encode real-world facts, which can effectively organize and represent knowledge so that it can be efficiently utilized in advanced reasoning-related applications, such as reasoning with commonsense knowledge [64, 73], reasoning with recommender system [63], and reasoning with logic relation inference [29].

A knowledge graph (KG), $G = (V, E)$ (e.g., WIKIDATA[2] and ConceptNet[3]) is a multi-relational graph. $V$ is the set of entity nodes, $E \subseteq V \times V$ is the set of edges that connect nodes in $V$. Knowledge graphs can provide structured information to better assist the reasoning tasks. In addition, the knowledge graph can be enriched with a set of observed facts/relations $O$ such that $E = V \times O \times V$. The observed fact is represented by a triplet $(v_i, o(v_i, v_j), v_j)$. For example, the triplet could be $v_i \rightarrow locates\_at \rightarrow v_j$ or $v_i \rightarrow requires \rightarrow v_j$ in ConceptNet. The explicit knowledge obtained from a knowledge graph can be categorized into two types: i.e., conceptual knowledge

---
[2]https://www.wikidata.org/wiki/Wikidata:Main_Page
[3]https://conceptnet.io/

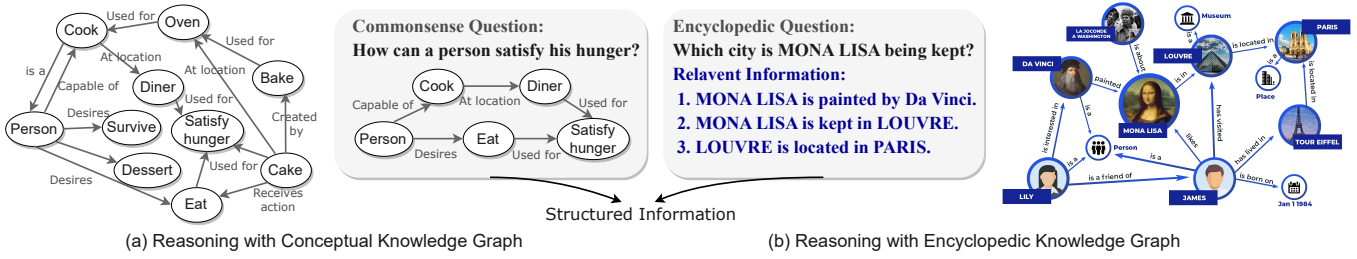

(a) Reasoning with Conceptual Knowledge Graph

(b) Reasoning with Encyclopedic Knowledge Graph

Fig. 3. The general schemes of reasoning with knowledge graphs[1].

and encyclopedic knowledge, each of which can be leveraged in specific reasoning tasks.

*4.1.1 Conceptual Knowledge Base.* A conceptual knowledge base [52] is usually an ontology-like knowledge graph, which is a collection of facts and information expressed in individuals, classes, and binary relations. For example, Figure 3 (a) visualizes the overall scheme of mining relevant information from a conceptual knowledge graph. Such relevant information could be individual paths that infer the reasoning procedure step-by-step or a subgraph that circulates the main context of the reasoning task and contains [64, 73]. For path-based knowledge retrieval, [23] have been retrieving reasoning paths from ConceptNet for commonsense reasoning tasks back to 2020, and [15] took a one-step further and leveraged reasoning paths retrieved from multiple heterogeneous knowledge graphs. Later on, [69] made the overall knowledge retrieval part explainable by a hierarchical path learner. For works that employed subgraph knowledge to aid reasoning, early work [23] proposed to generate answers/explanations for a given question by extracting a multi-hop relational subgraph from ConceptNet, which has triggered lots of work to extract multi-hop subgraphs as the external knowledge to solve the reasoning tasks. [64, 73] are two representative works that leverage subgraphs between entities from a logical question and entities from a pre-defined answer list as external knowledge in reasoning-related QA tasks.

*4.1.2 Encyclopedic Knowledge Base.* Encyclopedic knowledge bases [55] are centralized data repositories, including general knowledge about people, cities, countries, movies, organizations, etc. The form of an encyclopedic knowledge base varies from a collection of documents, a paragraph of descriptive text, and an encyclopedic knowledge graph. Encyclopedic knowledge graphs do not particularly emphasize the logical relationships between various entities, which is the primary distinction from conceptual knowledge graphs. As illustrated in Figure 3 (b), the essential goal of leveraging the Encyclopedic Knowledge Base as the knowledge source is effectively and efficiently retrieving relevant knowledge from the vast search space.

Earlier works [48, 71] tend to focus on learning network embeddings to conduct differentiable multi-hop knowledge extraction with considering a list of first-order logic rules. [74] later developed a learnable agent to conduct the knowledge retrieval in an end-to-end way by deep reinforcement learning. Other than mining from knowledge graphs, there are lots of datasets (e.g., SQUAD [46]) featuring multi-hop questions that are developed on encyclopedic knowledge (e.g., Wikipedia). Many works [62] are proposed to achieve competitive results with human annotators with various retrieval strategies.

## 4.2 Rule-based knowledge

The rule is used to explain the interaction between different objects in an environment. The relation $o(\cdot, \cdot)$, mentioned in the knowledge graph section, can also be considered as a predicate or first-order logic in the context of reasoning rules. For example, a ground definite clause can be written as:

$$o(v_i, v_j) \leftarrow o(v_i, v_p) \wedge o(v_p, v_q) \wedge o(v_q, v_j)$$

From this relation, we consider $v_p$ and $v_q$ as the mediators that can explain the relation between $v_i$ and $v_j$. Similar to conjunction ($\wedge$), a model can be improved with the knowledge of other different logical operations including disjunction ($\vee$), existence quantification ($\exists, \nexists$), universal quantification($\forall$), negation ($\neg$), etc. The Boolean operation over the atomic formulas makes the logic propositional. The possibilistic logic is defined as having a degree of truth between 0 and 1. Probabilistic logic is any sort of logic (predicate, propositional, etc.) that incorporates probability. Any logical conclusion that cannot be maintained in the face of fresh information is regarded as non-monotonic. In a broad sense, ontology can be seen as a rule that keeps track of the relationship between two entities. The knowledge of the rule can be implemented at the representation level or mapping level which we have discussed in this section.

*4.2.1 Rule Induced Representation.* In knowledge enhanced reasoning approach, both the symbolic and neural methods are combined to represent the hidden representation of input $X$ as $H = \phi(X; \{\theta, \Lambda\})$. where distributed representation is applied to design $\theta$ and rule induced another parameter $\lambda$ to improve the representation. The logic knowledge can be implemented with continuous relaxation in a deep learning network. The learning process allows jointly learning the weights of the deep learners and the meta-parameters controlling the high-level reasoning [40]. Representation is generally projected in Euclidean space. But geometric embedding can be one of the ways to attain the attributes of operators in the representation, such as hyperboloid embeddings in HypE [12] and cone embeddings in ConE [71]. In recursive reasoning network [20], specified and inferable information is embedded in the vector representation. Here vocabulary (Ontology) determines the recursive layers. Ontology can be applied with fuzzy reasoner to extract the most relevant features [54] for the representation. In a non-monotonic fashion,

---

[3]yashuseth.wordpress.com/2019/10/08/introduction-question-answering-knowledge-graphs-kgqa/

interactive learning updates the rules consistently to govern the representation [53].

*4.2.2 Rule Guided Inference.* Exclusive dependency on data-driven approaches for the hidden representation, $H' = \phi'(X; \theta)$, cannot capture the hidden rules properly. So, rule-guided inference can help to correct the reasoning at the downstream level. In that case, the rule-induced parameter $\lambda$ improves the mapping function $f(\cdot)$ where the output $Y$ can be mapped with input $X$ as $Y = f(\phi'(X; \theta); \{\omega, \lambda\})$. Here $\omega$ is the weight vector for the function and the $\lambda$ is the external rule-based constraint. In [65], authors used Markov Logic Networks (MLNs) to project First-Order Logic (FOL) as observed variables (symbolic knowledge) to correct error reasoning results. In the inference step, different types of rules including but not limited to ending anchored rule (EAR), cyclic anchored rule (CAR), bi-side ending anchored rule (bisEAR), the "rule of rule" (RofR) [72]; temporal rules [4], cyclic rules [37], etc. control the selection to assist the candidate inference.

## 5 APPLICATIONS

In this survey, we elaborate on a few prominent applications of reasoning that can be boosted by external knowledge. The eight branches of knowledge utilization (§2.2) can be used in each application based on the design choice.

### 5.1 Question Answering (QA)

Answering questions relies on real-world facts and relations, where the types of relationships could vary from single, multi-hop, or even complex-logical [70]. Typically, QA tasks are conventionally formulated as a structured query, and reasoning techniques are leveraged to find a hidden answer from the structure. For example: "Who is the great-aunt of Huey?" This question can be represented as a triplet $<s, P, o>$ and we need to find a subject ($s$) whether the object ($o$) is Huey and the predicate ($P$) is $GreatAuntOf$. For answering this question we can consider the following information:

- Function: $Female(Daphne)$,
- First-order-logic: $SiblingOf(Daphne, Quackmore)$, $GrandparentOf(Quackmore, Huey)$
- Clause: $GreatAuntOf(G, C) \leftarrow SiblingOf(G, I') \land GrandparentOf(I', C) \land Female(G)$

This will answer "Daphne is the great-aunt of Huey". In QA tasks, this different level of information can be used in the training or inference to deduct the answer. In RRN [20], the QA served as the outcome of ontological reasoning. In knowledge graphs, this triplet $<s, P, o>$ involves the task of entity prediction ($<?, P, o>$, $<s, P, ?>$) or relation prediction(relation type ($<s, ?, o>$), relation existence ($<s, P?, o>$)).

In Visual Question Answering (VQA), the goal is to provide a natural language answer for a given image, which involves collaborations of computer vision, NLP, and knowledge reasoning. The typical solution is to encode the image feature [1] and leverage LMs to predict an answer. Another way is to construct a directed scene graph from the image which is similar to a knowledge graph. Then the answer can be answered by multi-hop reasoning [28]. Some works also used spatial-aware image graphs [31]. All of these have been leveraging the help of external knowledge memory to understand the relations in images. The evaluation metrics consist of accuracy (overall, hits@n), consistency, validity, plausibility, grounding, and distribution scores.

In the above cases, the nature of the answer was a singleton. There is another form of QA which is scenario-based essay question answering (SEQA). The scenario can be described as $<Psg, Ant>$ where a passage ($Psg$) describes the scenario and image annotation ($Ant$) is given. Here the paragraph-long answer needs to be generated based on the scenario description ($<Psg, Ant>$) and the question context. In SR3 [9], authors used geographical scenarios to answer essay questions instead of multiple-choice questions in that domain. The evaluation metrics of these types of tasks can be ROUGE, BLEU.

### 5.2 Knowledge Completion

A large-scale knowledge base needs to be updated on a regular basis. Knowledge can be incomplete due to data collection methods, system inconsistency, human error, and missing options. This incomplete knowledge leads to data misrepresentation. To solve the problem, knowledge base completion is a very important application. The process is to complete triplets $<h, r, t>$ with head ($h$), tail ($t$), and relation ($r$). This leads to 10 different types of problem formulation including vector, the normal vector of hyperplane, embedding vector of relation, embedding vectors of head and tail, projection matrix, diagonal matrices, the dimensionality of an entity in embedding space, the dimensionality of relation in embedding space, the real part of a complex value and Hamilton product [67]. The inference engine can utilize the predefined rules to predict the missing links. These predefined rules can be single-type [11] multi-type [72], domain-specific ontology [26]. The inference process can also take the help of reinforcement learning to find the policy network [30]. In tensor-based rule mining, the interaction between different relations is captured by tensors. This mines different rules which can be utilized to complete the new triplets [60]. The evaluation metric for this knowledge completion uses Hits@n.

### 5.3 Explainable Recommendation

Recommendation systems must prioritize not only persuasion but also transparency, efficacy, and trustworthiness if they are to achieve user satisfaction ultimately. The reasoning focuses on users' behavior (e.g., action-based, user clicks), item features, and external knowledge base. The challenges involve generability, interpretation, and huge search space optimization. One way is to find the abstract of the users' behavior and then follow the explicit reasoning path for the explanation [61]. Case-based reasoning is also useful, especially to model human behavior in dealing with new problems [2]. Another way is to focus on the item features to find similar items for recommendation [38]. Using an external knowledge base can be used to comprehend the user, items, and user-item interaction properly [39]. Reasoning can also be applied to increase the level of intelligence from association to counterfactual. In that case, recommendations can be made under the "what-if" scenario [56].

### 5.4 Visual Reasoning

An image contains low and high-level content. Low-level content relates to object detection from the image. Deep learning methods have shown a huge improvement in this domain, yet reasoning can be helpful in object detection for domain adaptive issues [10], image quality issue [32], etc. High-level content is also important for various tasks in the visual domain specially for caption generation, labeling, image description, visual QA, etc. To capture the incomprehensible intention and the interaction of the objects' reasoning, there are works injecting external knowledge (e.g., ConceptNet) or internal knowledge (e.g., spatial position relationships [59]) into the encoder-decoder framework [21].

## 5.5 Others

Beyond the aforementioned applications, there are a lot of other applications for reasoning. Reasoning with heterogeneous graphs can improve the existing fact verification [57]. The conceptual graph can enhance the zero-shot and few-shot stance classification [35]. Reasoning by utilizing external knowledge varies widely based on domain-specific needs. Target-oriented Opinion Word Extraction (TOWE) is a new emerging subtask of Aspect Based Sentiment Analysis (ABSA) [14]. Existing domain knowledge can be used as collaborative memory for case-based reasoning [42]. When the domain knowledge is incomplete or the data is noisy; then the conclusion is derived defensibly, which can be further improved by non-monotonic logical reasoning based on further evidence [53]. In the absence of domain knowledge or previous explicit rule, the direction of the reasoning focused on converting data to explicit knowledge [36] or using evidential reasoning [33].

## 6 FUTURE DIRECTION

Current research trends in reasoning focused on multi-directional interaction between existing knowledge and the data. The concerns of the current researchers can lead future directions in the following ways:

*Brain-inspired AI..* In a recent white paper [66], researchers put more importance on understanding neural computation to improve the intelligence of AI. Compared to the animal brain, AI still faces challenges in three categories (the embodied Turing test, general intelligence, and sensorimotor capabilities). At a granular level, these categories consist of the challenges of energy efficiency, flexibility, interaction, adaptability, reasoning, abstract thought, making general inferences from sparse observation, combining streams of sensory information, and interaction with a new situation. To overcome the challenges, [66] suggested improving the research in behavioral flexibility (life-long memory, meta-learning, transfer learning, continual learning, self-supervised learning) and series challenges (iteratively optimizing, breaking to incremental challenges) by merging engineering and computation science, more theoretical and experimental research on neural computation and sharing platform for developing and testing virtual agents. External knowledge will benefit behavioral flexibility.

*The Ladder of Causation.* As a data-distribution-focused system, deep learning fails to understand the latent richness of human instruction. When it comes to converting a data-driven system to a causal relation-driven system, then most of the approaches are still stuck at the beginning level of the causation ladder. Pearl defined the three levels of the ladder consists of association ($P(Y|X)$), intervention ($P(Y|do(X))$), and counterfactual ($P(Y_{x=0} = 0|X = 1, Y = 1)$) [45]. Knowledge-enhanced reasoning can help the data-driven process to the upper rung, but still it has the limitation of rule-based learning. In the future, reasoning can be enriched by finding the hidden confounder and the mediators in the data. These will improve the current induction reasoning ("Evidence-to-Hypothesis") to a great extent.

*Hybrid (Switchable) System.* The deep learning process is analogous to Mysterious-Consciousness (M-consciousness) or System-I [5]. Current reasoning research is focusing on building Information-Consciousness (I-consciousness) or System II, but an important architectural feature should also be involved in this development process and that is switchability. Human has the ability to switch from System-I to System-II, single agent to multi-agent. Whether is it possible to make a generic system that can make conscious decisions about which system to use? Do we need any other knowledge to achieve that? This can direct the development of the hybrid system to multi-source knowledge reasoning which can conduct dynamic reasoning and zero-shot reasoning efficiently.

## 7 CONCLUSION

The development of reasoning-capable AGI is a multifaceted study topic where external knowledge-augmented research has justifiably gained the lion's share of the credit. To meet the ongoing demand for a properly categorized overview of this research topic, we have reviewed the most recent developments in this field and provided a comprehensive categorization of the knowledge utilization. We also provided the application perspective in the hope that it will assist the reader in understanding the current popular applications and how to solve the tasks with knowledge-enhanced reasoning. We concluded our survey by focusing on three topics that have recently emerged in the spotlight of AGI.